\documentclass[letterpaper, 10 pt, conference]{ieeeconf}  % Comment this line out
\IEEEoverridecommandlockouts                              % This command is only
\usepackage{times} % assumes new font selection scheme installed
\usepackage{amsmath} % assumes amsmath package installed
\usepackage{amssymb}  % assumes amsmath package installed

\usepackage[utf8]{inputenc}
\usepackage{booktabs}
\usepackage{multirow}
\usepackage{subcaption} %-- This makes table captions normal looking and smaller
\usepackage{array}
\usepackage{xspace}
\usepackage{flushend}
\usepackage{algorithm}
\usepackage{algorithmicx}
\usepackage{tabularx}
\usepackage{graphicx}
\usepackage{placeins}
\usepackage[noend]{algpseudocode}
\usepackage{hyperref}
\hypersetup{urlcolor=blue, colorlinks=true}
\usepackage[colorinlistoftodos,prependcaption,textsize=tiny]{todonotes}

\newcommand{\shohin}[1]{\textcolor{black}{#1}}

\newcommand{\revii}[1]{\textcolor{black}{#1}}
\newcommand{\reviii}[1]{\textcolor{black}{#1}}
\newcommand{\state}{\ensuremath{\mathbf{q}}\xspace}
\newcommand{\obs}{\ensuremath{\mathbf{o}}\xspace}

\newcommand{\skill}{\ensuremath{\mathbf{s}}\xspace}
\newcommand{\Skillset}{\ensuremath{\mathcal{S}}\xspace}

\newcommand{\predicate}{\ensuremath{p}\xspace}
\newcommand{\Predicateset}{\ensuremath{\mathcal{P}}\xspace}

\newcommand{\plan}{\ensuremath{P}\xspace}
\newcommand{\Goalcond}{\ensuremath{L_G}\xspace}
\newcommand{\Effect}{\ensuremath{L_E}\xspace}
\newcommand{\Precond}{\ensuremath{L_P}\xspace}

\newcommand{\blockred}{\textsc{BlockRed}\xspace}
\newcommand{\blockgreen}{\textsc{BlockGreen}\xspace}
\newcommand{\blockblue}{\textsc{BlockBlue}\xspace}
\newcommand{\blockyellow}{\textsc{BlockYellow}\xspace}

\newcommand{\reachtable}{\textsc{ReachOnTable}}
\newcommand{\reachblock}{\textsc{ReachOnTower}}
\newcommand{\stack}{\textsc{Stack}}
\newcommand{\unstack}{\textsc{Unstack}}
\newcommand{\pull}{\textsc{Pull}}
\newcommand{\singulate}{\textsc{Singulate}}

\newcommand{\on}{\textsc{On}}
\newcommand{\inhand}{\textsc{InHand}}
\newcommand{\ontop}{\textsc{OnTop}}
\newcommand{\inworkspace}{\textsc{InWorkspace}}
\newcommand{\close}{\textsc{Close}}

\usepackage[font=scriptsize,labelfont=bf]{caption}
\begin{document}
%\title{Learning Visuomotor Policies for Real-World Task Planning and Execution}
\title{Reactive Long Horizon Task Execution via \\ Visual Skill and Precondition Models}

\author{Shohin Mukherjee$^{1,2}$, Chris Paxton$^{1}$, Arsalan Mousavian$^{1}$, Adam Fishman$^{1, 3}$, Maxim Likhachev$^{2}$, Dieter Fox$^{1, 3}$
% \thanks{$^{1}$ NVIDIA, USA
%         {\tt\small \{cpaxton, amousavian, dieterf\}@nvidia.com}}%
% \thanks{$^{2}$ Carnegie Mellon University, USA
%         {\tt\small \{shohinm, mlikhach\}@andrew.cmu.edu}}
% \thanks{$^{3}$ University of Washington, USA 
%         {\tt\small afishman@cs.washington.edu}}
\thanks{$^{1}$ NVIDIA, USA\newline
        {\tt\small \{cpaxton, amousavian, dieterf\}@nvidia.com}}%
\thanks{$^{2}$ Carnegie Mellon University, USA\newline
        {\tt\small \{shohinm, mlikhach\}@andrew.cmu.edu}}
\thanks{$^{3}$ University of Washington, USA\newline
        {\tt\small afishman@cs.washington.edu}}
}
\maketitle

\begin{abstract}
Zero-shot execution of unseen robotic tasks is important to allowing robots to perform a wide variety of tasks in human environments, but collecting the amounts of data necessary to train end-to-end policies in the real-world is often infeasible. We describe an approach for sim-to-real training that can accomplish unseen robotic tasks using models learned in simulation to ground components of a simple task planner. We learn a library of parameterized skills, along with a set of predicates-based preconditions and termination conditions, entirely in simulation. We explore a block-stacking task because it has a clear structure, where multiple skills must be chained together, but our methods are applicable to a wide range of other problems and domains, and can transfer from simulation to the real-world with no fine tuning. The system is able to recognize failures and accomplish long-horizon tasks from perceptual input, which is critical for real-world execution. We evaluate our proposed approach in both simulation and in the real-world, showing an increase in success rate from 91.6\% to 98\% in simulation and from 10\% to 80\% success rate in the real-world as compared with naive baselines. For experiment videos including both real-world and simulation, see: \url{https://www.youtube.com/playlist?list=PL-oD0xHUngeLfQmpngYkGFZarstfPOXqX}
\end{abstract}

\FloatBarrier
\section{Introduction}
\label{sec:introduction}
% In order to reliably complete complex tasks without human aid, robots must be able to reason about unseen circumstances, recognize failures, and generalize to new environments. While robot-learning can excel at a wide variety of specific manipulation tasks~\cite{rajeswaran2017learning}, there is still a question as to how to design systems that can scale up in the real-world to accomplish long-horizon goals.

Robot learning has shown a great ability to learn skills from perception data without requiring a large amount of human intervention~\cite{rajeswaran2017learning,zeng2020tossingbot}. A number of works have looked at learning higher level policies capable of generalizing to different tasks~\cite{xu2018neural,huang2019neural,hundt2020good}, but these methods are all very data intensive. As a result, sim-to-real transfer has become an increasingly important research area~\cite{tobin2017domain,chebotar2019closing,blukis2019learning}. High-fidelity simulators enable the collection of a large amount of data quickly and cheaply, which can then be used to train skills.
Though training policies in simulation solves the data gathering  problem, these policies perform poorly in the real-world because of the \textit{reality gap}.
Techniques like domain randomization~\cite{tobin2017domain,tremblay2018deep} have been used to allow policies trained on visual perception to transfer from simulation to the real-world. In addition, depth-based models such as UOIS~\cite{xie2020best} or 6-DOF GraspNet~\cite{mousavian20196,murali20206} transfer better than RGB-based models, with little to no real-world fine tuning necessary.
Despite these strategies, real-world execution of long horizon tasks remain a challenge because as the time horizon of execution increases, so does the likelihood of the policy failing. An alternate solution is to design systems that have the ability to detect and react to failures.

To build such a reactive system, we can draw inspiration from task planning~\cite{paxton2019representing} and task and motion planning~\cite{toussaint2015logic,garrett2018ffrob,garrett2020online}. These methods can do an excellent job at solving new tasks in known, modeled environments, and can even achieve a good amount of reactivity and robustness in these environments using various techniques~\cite{paxton2019representing,garrett2020online}. Task planners define a planning domain that lists a set of preconditions and effects corresponding to each action. These preconditions and effects are based on predicates that define \emph{symbolic relationships} between entities in the world. Evaluation of these predicates enables the detection of failures so that a replanning request can be triggered to handle these failures. However it is unclear how to ground these relationships in the real-world where the true state of the entities is unobservable.

\begin{figure}[bt]
\includegraphics[width=\columnwidth]{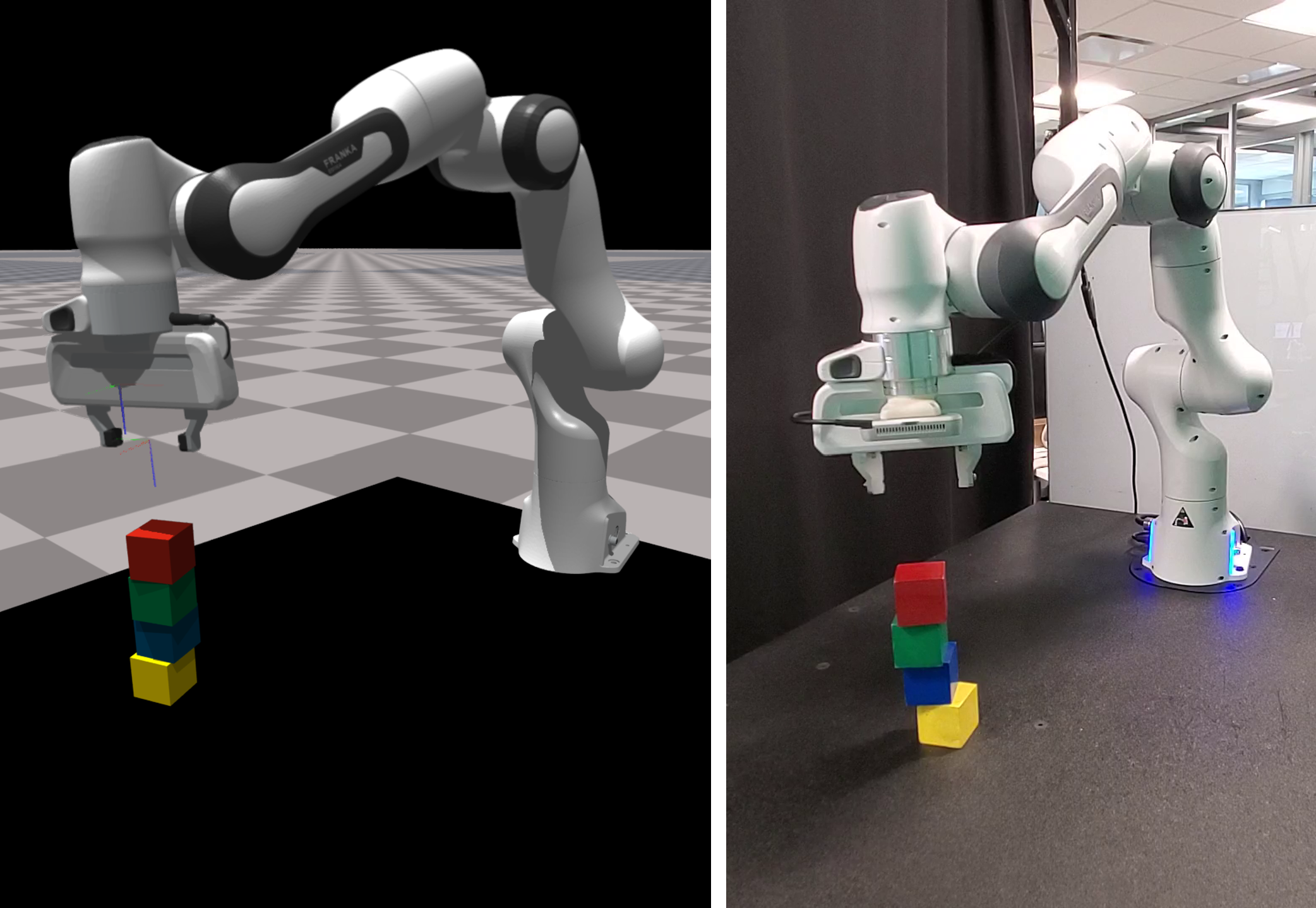}
\caption{Our system performing a stacking task both in simulation (left) and the real-world (right). Our framework integrates a mixture of learned and manually-designed manipulation policies, unified by a task planning framework, in order to achieve robust, reactive execution both in simulation and in the real-world, even in the presence of much higher levels of sensor noise.}
% \vskip -0.5cm
\end{figure}

% \textbf{The main objective of our work is the following:} \shohin{We want 
Our objective is to learn perceptual models for these predicates along with manipulation skills, purely from simulation data. We use these models
%together via a task-planning framework
to build a reactive system that can reliably execute long-horizon manipulation tasks. Specifically, we look at a stacking task similar to those explored in prior work~\cite{xu2018neural,hundt2020good}.
Our key insight is that models for the predicates will generalize better to the real-world than the learned policies, allowing us to use reactivity and failure recovery to compensate for the loss of precision that occurs due to using policies trained purely in simulation in the real-world. 

\begin{figure*}[bt]
    \centering
    \includegraphics[width=0.98\textwidth]{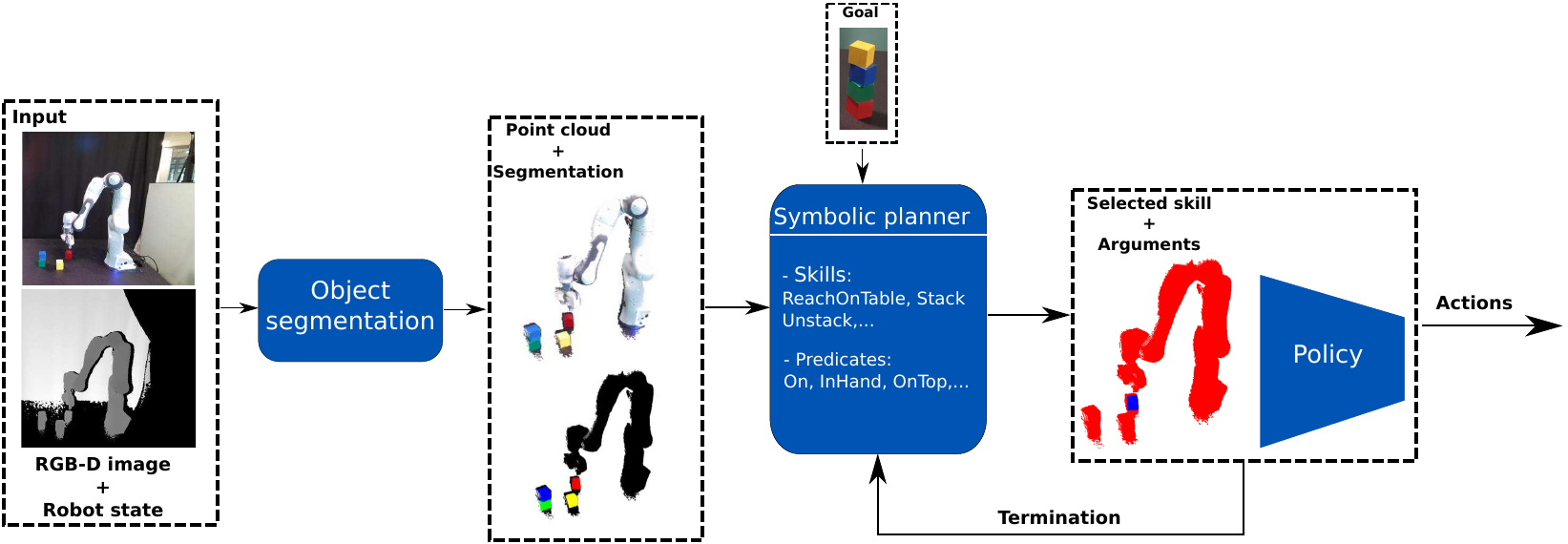}
    \caption{Overview of our method. The input is the RGB-D image of the scene and robot joint configuration. Objects in the scene are segmented and this semantic information is passed to the symbolic planner. The planner uses the predicate models to construct a symbolic state and generates a plan to achieve the target configuration from the current state. Each step of the plan is a high-level skill and its corresponding arguments. Each skill has a learned policy model that predicts the low-level joint command and a termination model that determines whether or not the skill is finished executing. Upon termination of a skill, we determine the next skill to execute based on the current logical state, and replan only if our current plan contains no executable skills.}
    \label{fig:overview}
    % \vskip -0.5cm
\end{figure*}

%\textbf{Our work differs from prior work on robot-learning in the following key ways:}
%\shohin{Firstly, since it is based on task planning, 
This approach gives us the generalization power to successfully achieve tasks not present during training as long as the necessary predicates and skills have been learned.
The skills are trained to accomplish a low-level objective and serve as local policies that are distilled into a task plan that serves as the high-level policy.
%For any new task, the pre-existing models do not need to be retrained; only the new set of skills that are required to accomplish a new (possibly harder) task need to be trained and added to the task model.
For any new (possibly harder) task, only the set of new, additional skills must be trained and added to the task model.
This allows for a ``plug-and-play'' approach as the robot builds up a library of skills to accomplish tasks with increasing complexity without having to re-invent the wheel.
Secondly, we rely on failure detection and recovery using learned models for the predicates; as described in Section~\ref{sec:methods/learning}, these models are binary classifiers and are therefore easier to train and transfer better to the real-world than the policy models .

To summarize, our contributions are:
\begin{enumerate}
    \item A plug-and-play approach to build up a library of skills to accomplish long-horizon manipulation tasks;
    \item a novel visuomotor policy learning approach from demonstration data collected from a planner using a PointNet++ based model that transfers from simulation to the real-world in one-shot;
    \item and a highly reactive system that detects and elegantly recovers from execution failures in the real-world, using a task planning framework that uses learned predicate models.
\end{enumerate}

% As it is based on task planning, this gives us generalization to held-out robot tasks ``for free'' so long as the necessary predicates and actions have been learned.

% \FloatBarrier
\section{Related Work}

%\begin{figure*}[bt]
%    \centering
%    \includegraphics[width=0.95\textwidth]{figures/overview.eps}
%    \caption{Overview of our method. The input is the RGB-D image of the scene and robot joint configuration. Objects in the scene are segmented into different blocks. The symbolic planner takes the information from segmented point cloud, evaluates predicates, and plans to achieve the target tower configuration. Each step of the plan is a high level skill and its corresponding arguments. Each skill has a policy that predicts the action that needs to be taken and whether that skill is finished executing or not. Upon termination of skill, the symbolic planner re-plans to achieve the target configuration.}
%    \label{fig:overview}
%\end{figure*}

%\textbf{Hierarchial policy learning:}
There is a large and growing body of research on hierarchical robot policy learning. This involves  hierarchical decomposition of a complex manipulation task into low-level tasks for which skills can be learned \cite{kroemer2015towards} with hierarchical reinforcement learning and the options framework \cite{konidaris2009skill}. Decomposing tasks has the advantage that the individual component skills can be learned more efficiently because each skill is of shorter horizon \cite{kroemer2019review}. This work is inspired by this philosophy, but it takes a task-planning approach to chaining of the component skills instead of a high-level policy.

%\textbf{Task and motion planning:}
A different approach for solving complex multi-step robot manipulation problems is Task and Motion Planning (TAMP). TAMP has previously been extended to stochastic actions and partially-observable environments~\cite{garrett2020online}.
Recently, there has been more interest in using learning to improve TAMP as well. Driess et al.~\cite{driess2020deep} learn a visuomotor policy for guiding exploration in task and motion planning. Kim et al.~\cite{kim2019learning} learned a score-space representation to guide task and motion planning based on constraints, and Paxton et al.~\cite{paxton2017combining} used Q-learning to guide tree search over a set of black box (learned) policies for autonomous driving.

%\textbf{Representation learning:}
Another body of work looks at learning representations that can be used for planning. Regression planning networks plan symbolic actions back in time from some future goal state~\cite{xu2019regression}. Visual robot task planning plans in a learned latent space, captured via an autoencoder~\cite{paxton2019visual}. Huang et al. propose a continuous relaxation for symbolc planning~\cite{huang2019continuous}, which lets them plan over probabilities for learned predicates instead of exact predicate values.
In~\cite{manuelli2020keypoints}, the authors propose using keypoints in RGBD images to learn a dynamics model in the context of model-free reinforcement learning.
Broadly-Exploring Local-Policy Trees~\cite{ichter2020broadly} use learned local policies to handle exploration through a learned latent space.

In order to get an extensible, general planning approach capable of solving unseen tasks, we want a framework that is agnostic to whether or not individual components are learned or not. This is not necessarily a new idea; task and motion planners such as PDDLStream~\cite{garrett2020pddlstream} use a sequence of ``black box'' planners, which could theoretically be learned. Wang et al.~\cite{wang2018active} learn predicates for use in task and motion planning, and Huang et al.~\cite{huang2019continuous} also learn predicate models. We specifically look to learn these for use on commodity sensors, trained in simulation and applied in the real world.

The approach described in~\cite{paxton2019representing,kase2020transferable} has the advantage of letting us combine black-box classifiers and policies, and potentially combining them with an off-the-shelf task planner like FastDownward~\cite{helmert2006fast}. These define actions as PDDL-style operators, with preconditions and effects surrounding a black-box policy potentially operating on raw sensor data. However, Kase et al.~\cite{kase2020transferable} learns a single monolithic policy, and does not explicitly learn a model of termination conditions (effects), which limits transfer and generalization and does not allow us to learn separate specialized skills or easily extend the model.
%In our work we train separate models for different classifiers, though a hybrid approach is possible.

In this work, we use the approach proposed in~\cite{yan2020close} for sim-to-real, wherein segmentation masks are used as the interface between perception and control in order to close the sim-real gap.
%Since semantic information is freely available in simulation, policies can be trained entirely on this with a smaller network, and directly transferred to the real-world.
Other works have shown that segmentation masks get better generalization when learning policies, e.g.~\cite{roh2020conditional}. Our policies are based on PointNet++, which has previously been shown to transfer well from simulation to real sensors~\cite{mousavian20196,murali20206}.

% \FloatBarrier
\section{Methods}
%\begin{figure*}[!htb]
%    \centering
%    \includegraphics[width=0.8\textwidth]{figures/overview.eps}
%    \caption{Overview of our method. The input is the RGB-D image of the scene and robot joint configuration. Objects in the scene are segmented into different blocks. The symbolic planner takes the information from segmented point cloud, evaluates predicates, and plans to achieve the target tower configuration. Each step of the plan is a high level skill and its corresponding arguments. Each skill has a policy that predicts the action that needs to be taken and whether that skill is finished executing or not. Upon termination of skill, the symbolic planner re-plans to achieve the target configuration.}
%    \label{fig:overview}
%\end{figure*}

%We first give an overview of our approach before delving into the finer details.
Our method is a hybrid of classical symbolic planning and recent deep learning techniques, which allows us to combine both learned and manually-designed predicates and policies into the same framework. We learn a set of \emph{skills} \Skillset, over which we can use a symbolic planner to find sequences of skills that can be applied in new environments to accomplish unseen tasks.
% More concretely, our approach consists of a symbolic planner and a set of skills.
%The predicates are used to define the high-level state of the world. The symbolic planner plans a series of high-level actions to achieve a goal-state.
Formally, each skill is represented as a set of four functions:
\[
   \skill = \left( \Precond, \Effect, \pi, f_T\right)
\]
\noindent where \Precond is the set of logical \emph{preconditions}, \Effect is the set of expected logical \emph{effects}, $\pi$ is a visuomotor policy, and $f_T$ is the \emph{termination} condition for that policy.

All of these are parameterized by some number of optional arguments, representing the entities on which the function is acting. For example, in a reaching skill, there is one argument, which is the object to be grasped; for a stacking skill, there would be two arguments, for the object in hand and the object upon which it is being stacked.

\Precond and \Effect are the logical preconditions and effects associated with the skill \skill. These govern when it can be used, and which properties of the world should change, and, how they change after the skill's termination function $f_T$ evaluates to true. These are essentially sets of \emph{predicates} $\predicate \in \Predicateset$ for which we learn models, although these could be learned in a variety of different ways based on different data or could be manually-designed if necessary.

%
%The policy $\pi$ captures the action that must be performed (e.g. reaching, stacking, unstacking) along with its arguments: the specific entities that the skill is acting on.
%These arguments specify relevant information for the selected skill such as which block to pick up, which object to stack a block on and so on.

Each skill is associated with a termination function $f_T$ that predicts a termination probability that informs the execution algorithm that the skill execution is complete and it can move to the skill corresponding to the next  high-level action in the plan or re-plan. The plan is also comprised of a set of preconditions (based on the predicates) associated with each skill. A skill \skill can only be executed if all its preconditions \Precond are satisfied by the current observation. The execution algorithm evaluates the preconditions and adjusts the plans accordingly.

%The key ingredients of our approach are a set of skills \Skillset along with their termination triggers, a set of predicates \Predicateset and a task planner. 

Each policy $\pi$ is either a learned model or manually-designed. Here we focus on the learned policies included in our framework. % in case it can be trivially done. 
The learned policies are trained from \shohin{data generated by execution of an expert policy in a simulator using imitation learning. The expertise comes from the fact that the planner has access to privileged information from the simulator.} \shohin{During execution the trained policies operate purely on perceptual input in the form of point cloud from an RGB-D camera and readily available joint angles from the robot's encoders.} We use NVIDIA Isaac Gym~\cite{liang2018gpu} as the simulation platform because of its high-fidelity and ability to run parallel environments using GPU acceleration. To facilitate imitation learning, it is desired that the expert policy is consistent, i.e. given the same goal and initial configuration, the expert will predict the same trajectory to the goal. Therefore, we use a search-based planner known as ARA*\cite{likhachev2004ara} that can compute deterministic plans with bounded sub-optimality for each pair of start and goal states. Fig.~\ref{fig:overview} shows the overview of our approach.

%Unlike sampling-based planners, search-based planners are deterministic which ensures that given a fixed start state, the expert will always predict the same trajectory to the goal -- a property that is key for imitation learning.

% Given an instance of a scene and a set of goal conditions \Goalcond, the task planner uses the predicates in \Predicateset to predict a logical state on the initial observation $\obs_{start}$ to generate a task plan using STRIPS-style symbolic planning. The task plan comprises of a list of grounded actions each of which correspond to a grounded skill along with a set of intrinsic and extrinsic termination preconditions \todo[author=Chris]{please explain the difference} \arsalan{definitely agree with Chris here}. A skill \skill can only be executed if all it's preconditions are satisfied by in the current observation $\Precond^\skill$. The task plan is then executed on the real robot to achieve successful one-shot sim-2-real transfer.

\subsection{Expert Data Generation}

\begin{figure}[tb]
    \centering
    \includegraphics[width=\columnwidth]{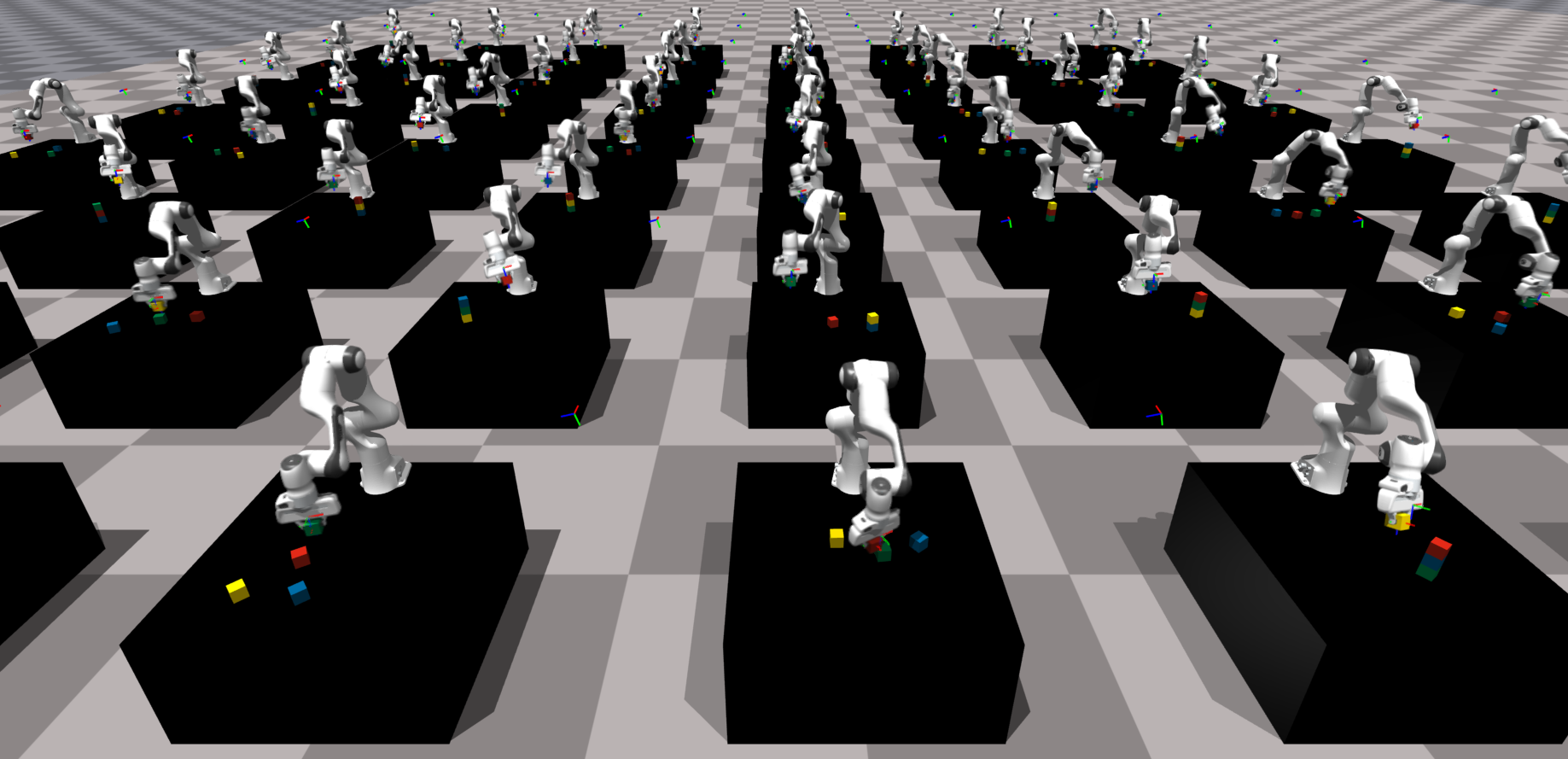}
    %\caption{Trajectories generated by expert for \stack being executed in parallel in simulation.}
    % \vskip 0.2cm
    \includegraphics[width=1.0\columnwidth]{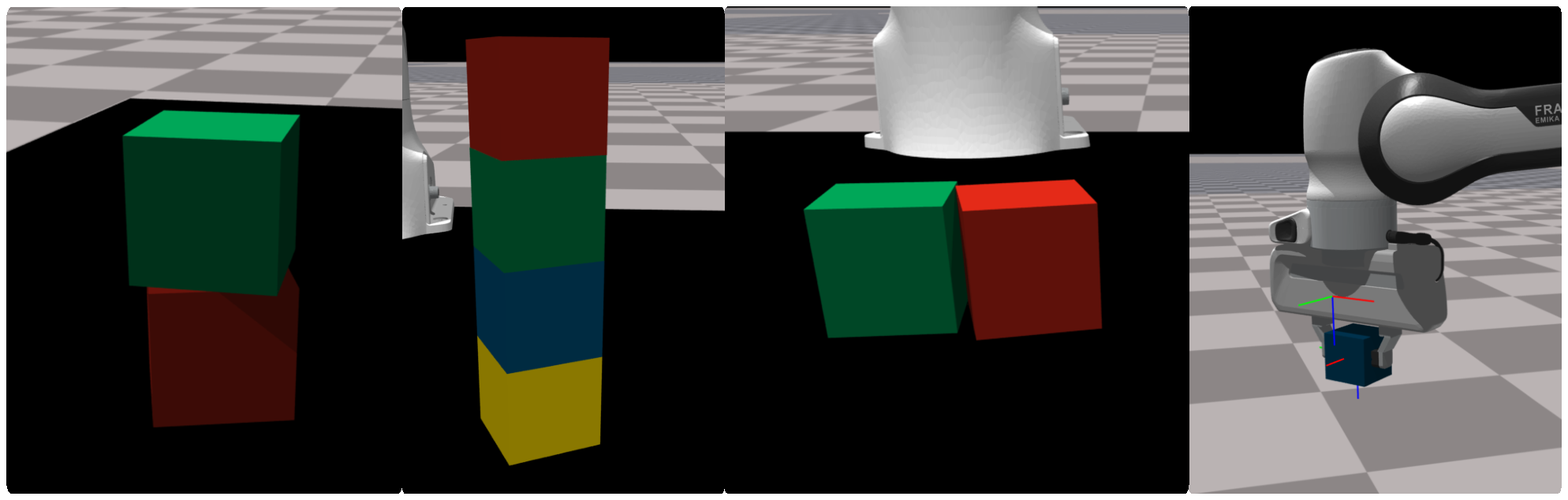}
    \caption{Top: Expert data collection in simulation for the \stack~skill. We run 64 parallel agents, allowing us to collect 15,000 trials in 3 hours. Bottom: examples of ``true'' predicates from simulation data collection.
    From left to right: \on(\blockgreen, \blockred), \ontop(\blockred), 
    \close(\blockred, \blockgreen) and \inhand(\blockblue).
    }
    \label{fig:sim_data_collection}
\end{figure}

The expert planner uses a model of the scene to generate a set of feasible trajectories. The trajectories correspond to the skill that is being learned. For example, for the \reachtable\xspace action, we generate a set of collision-free trajectories from a random start configuration of the robot to a goal configuration which would allow the robot to grasp a particular block. The generated trajectories are then executed in parallel in the simulation, and the robot joint states along with the camera point cloud are recorded at a predefined frequency as shown in Fig.~\ref{fig:sim_data_collection}. We want the skills and the corresponding termination triggers to be particularly accurate close to the goal. Therefore, we densely sample states around the last state of the expert trajectories and append these to the trajectories in the data set. These states increase the precision of the skill models and serve as hard-negatives for the termination models.

\subsection{Skills and Predicates Learning}
\label{sec:methods/learning}
\begin{figure*}[tb]
    \centering
    \includegraphics[width=\textwidth]{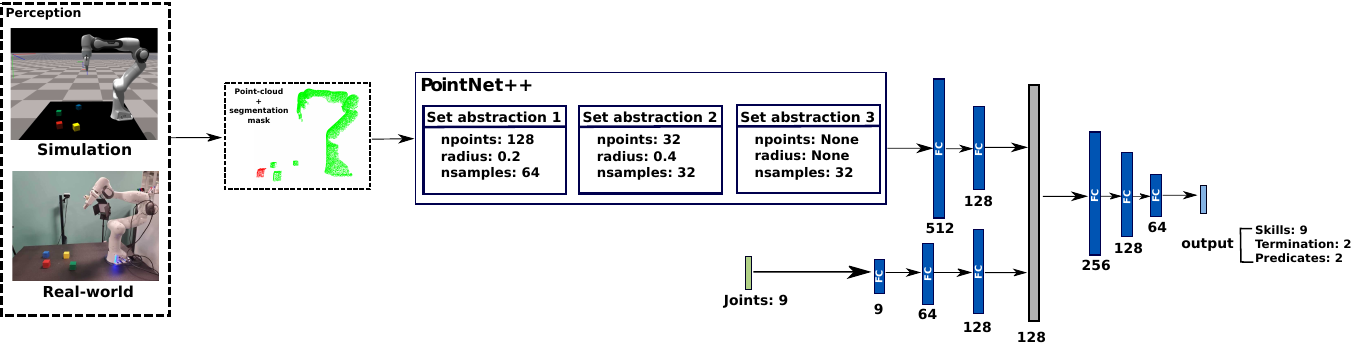}
    \caption{Overview of the model architecture used for skill learning and for modeling predicates. The model is based on Pointnet++ architecture. The input to the model is the point cloud of the scene and features associated with each point. The point features have extra information that indicate the arguments for the skill/predicate. The model processes this information and predict the final joint configuration and termination probability for skills and probability for predicates.}
    \label{fig:network}
    % \vskip -0.2cm
\end{figure*}

We will refer to the state of the robot's joints as \state and the camera observation will be referred to as \obs. Each of the actions in the task plan correspond to a skill $\skill \in \Skillset$. Each skill is comprised of a visuomotor policy. The policy predicts the next state \state for the robot to go to in joint-space given an observation \obs. Given a sequence of states and observations in an expert trajectory $\{\state,\obs\}_{t=0}^T$, the skill could be trained to predict a state $\state_{t'}$ for an observation $\obs_t$ s.t. $t'>t$. In this work, the skill is trained to predict the final state in the expert trajectories, so $t'=T$. Each skill's associated termination condition $f_T$ predicts the probability of the skill terminating given an observation: $f_T: \obs \rightarrow [0, 1]$. \shohin{The termination models are trained as binary classifiers}. The final observations in the expert trajectories $\obs_T$ are positive examples for the termination models $f_T$, and the rest of the observations  $\obs_{t<T}$ are negative examples.  

We also train a set of predicates which are used to construct a symbolic state of the world for task planning and recovering from failures. The predicates have the same model architecture as the termination conditions for the skills and they predict the probability of the predicate being satisfied given an observation . The learned predicates are trained on the same data that is used to train the skills. In simulation we have access to all geometric information, so we have access to the ground-truth values for all predicates given an observation. \shohin{The learned models can then be used to predict the values for the corresponding predicates directly from perceptual input during execution.} Table~\ref{tab:skills_predicates} lists all the skills and predicates. Fig.~\ref{fig:sim_data_collection} (bottom) shows examples of scenes when these predicates are satisfied.

\begin{table}[bt]
\resizebox{\columnwidth}{!}{%
\begin{tabular}{cllc}
\toprule
\multicolumn{2}{c}{}                    & Description                             & Learned/Manual \\\midrule
%\multirow{6}{*}{
\parbox[t]{2mm}{\multirow{6}{*}{\rotatebox[origin=c]{90}{\textbf{Skills}}}}
%}
&\reachtable(X)  & Grab block X on table                   & Learned        \\
&\reachblock(X)  & Grab block X on another block           & Learned        \\
&\stack(X,Y)     & Stack block X on Y                      & Learned        \\
&\unstack(X)     & Unstack block X                         & Manual         \\
&\pull(X)        & Pull block X to workspace               & Manual         \\
&\singulate(X,Y) & Separate blocks X and Y from each other & Manual        \\
\midrule
%\multirow{5}{*}{Predicates}
\parbox[t]{2mm}{\multirow{5}{*}{\rotatebox[origin=c]{90}{\textbf{Predicates}}}}
&\on(X,Y)        & Block X is on block Y                          & Learned        \\
&\inhand(X)      & Block X is in the robot's gripper              & Learned        \\
&\ontop(X)       & There is no other block on top of block X      & Learned        \\
&\inworkspace(X) & Block X is in the robot's workspace            & Manual        \\
&\close(X,Y)     & Blocks X and Y are too close to each other     & Manual         \\

\bottomrule
\end{tabular}%
}
\caption{Learned and manual skills and predicates. We present a framework where we can combine skills from different sources; in our case study, certain skills and predicates were designed manually.}
\label{tab:skills_predicates}
% \vskip -0.3cm
\end{table}

\begin{figure}[!htb]
    \centering
     \includegraphics[width=0.8\columnwidth]{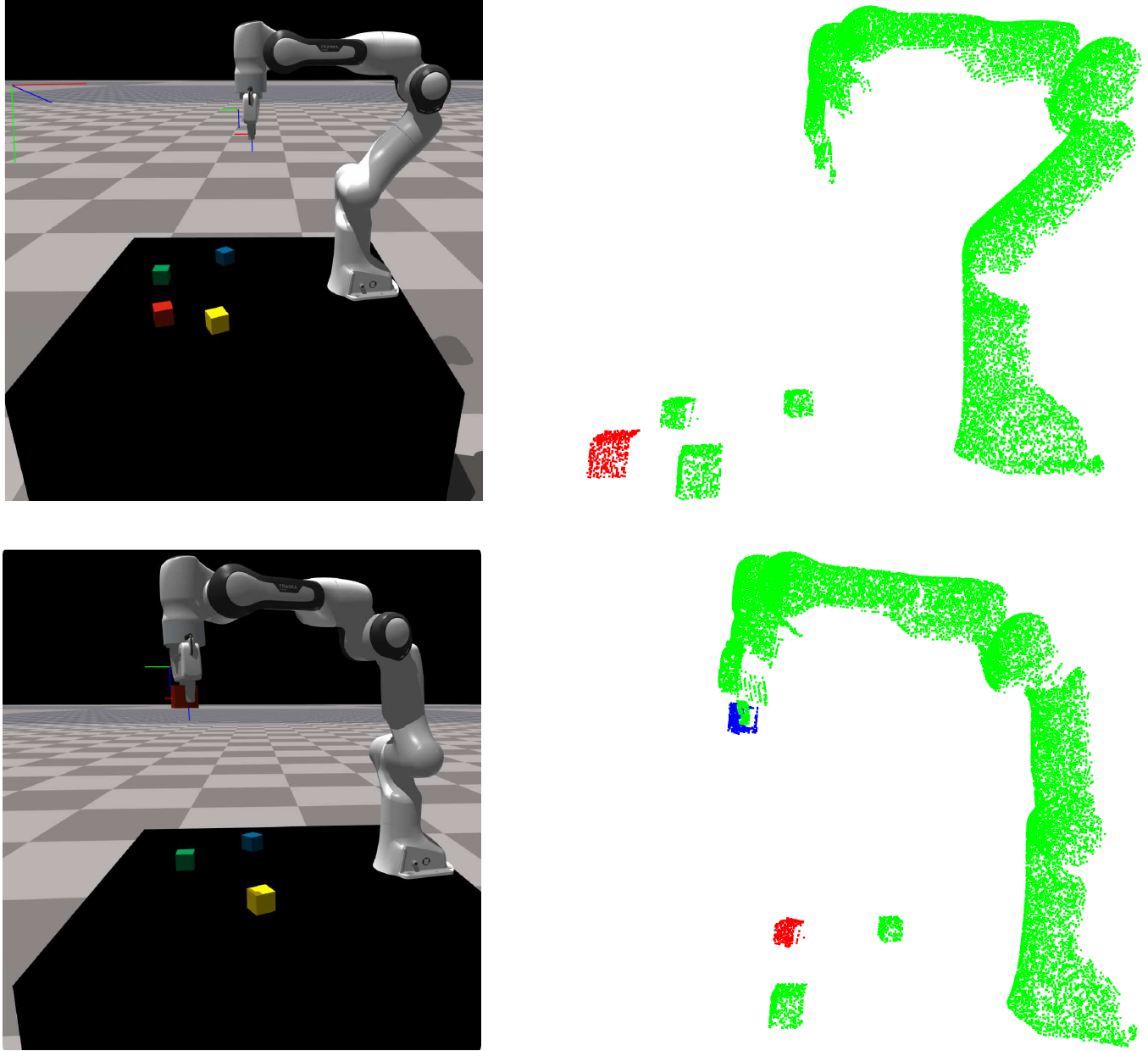}
    \caption{Point cloud with mask for \reachblock(\blockred) (top) and \stack(\blockred, \blockgreen)  (bottom). Note that the color in the point cloud corresponds to the binary masks and does not correspond to the color of the block.}
    \label{fig:skill_pc}
\end{figure}

For the learned skills, termination conditions and the predicates the model architecture is shown in Fig.~\ref{fig:network}. The point cloud from the camera is first processed to append a binary mask on the points that belong to an object of interest to a particular model as shown in Fig.~\ref{fig:skill_pc}. For example, for the skill \reachtable$(\blockred)$ or the predicate \inhand$(\blockred)$, a binary mask is appended to the points that belong to \blockred. Similarly, for the skill \stack$(\blockred, \blockgreen)$ or predicate \on$(\blockred, \blockgreen)$, masks are appended to blocks \blockred and \blockgreen. We then use PointNet++~\cite{qi2017pointnet++} to extract the global features for the processed point cloud. The point cloud features are appended with the features extracted from the joint states and then used to predict either the joint targets, in case of the skills, or a probability in case of the termination conditions and predicates. The skills are trained with a linear combination of a joint-space and an operational-space loss which are explained below, whereas the termination conditions and the predicates are trained with a cross-entropy loss.

\textbf{Joint-space loss}: The joint-space loss is simply an MSE loss between the predicted joint states and the final joint states of the expert trajectory. 

\textbf{Operational-space loss}: This loss is used to minimize loss to the final Cartesian end-effector position. We represent the end-effector position with the two points, one of each on the x- and y- axes of the end-effector coordinate frame, respectively. These two points uniquely determine the robot's end-effector frame. Using forward kinematics, we map \state for both the prediction and the expert onto these points. We then take the Euclidean distance between the corresponding points on the expert and the prediction. Choosing which specific points along the axes is a matter of tuning, as their distance from the origin essentially trades off between rotation and translational penalties.

\subsection{Task Planning and Execution}
\begin{figure*}[tb]
    \centering
    \includegraphics[width=1.0\textwidth]{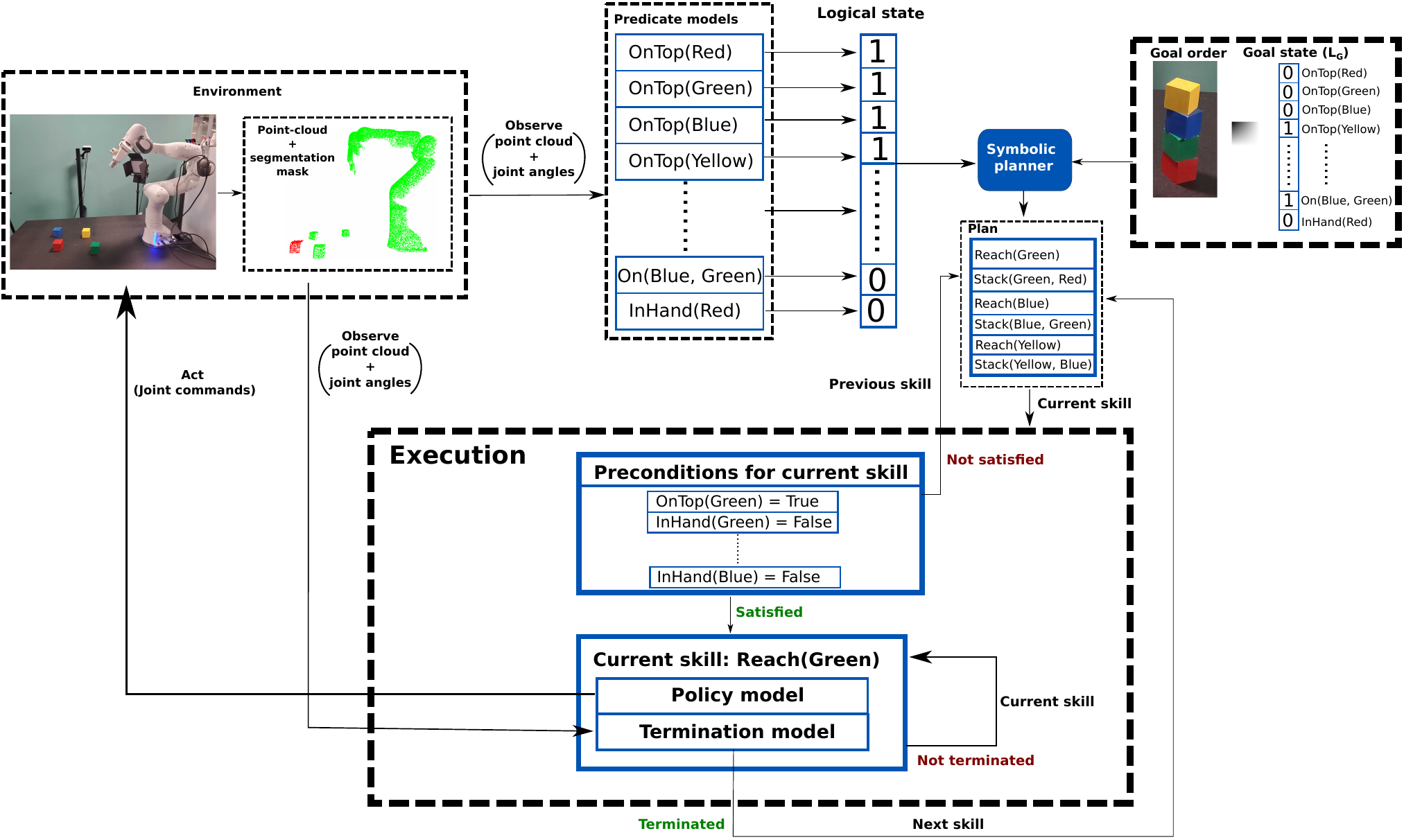}
    \caption{\shohin{Full system overview, showing coupling of perception with high-level task planning and execution.
    The trained predicate models receive two perceptual inputs: (1) a point cloud of the scene with a segmentation mask generated by an off-the-shelf segmentation algorithm, and (2) the robot's joint positions. These are used to generate a binary vector representing the logical state for the scene. The symbolic task planner uses a pre-defined task-model to generate a high-level task plan, which is a sequence of skills along with their arguments and a set of preconditions for each skill (based on the predicate models). Each skill in the task plan has a corresponding policy model and a termination model. For the current skill in the plan, the preconditions are evaluated. If they are not satisfied, the control goes to the previous skill in the plan. If they are satisfied, the current skill can be executed. For the skill, the policy model sends out joint commands to the robot to follow. The termination model receives the perceptual input from the environment and predicts if the skill has terminated. The current skill stays active until termination. Upon termination, the control goes to the next skill in the the plan for execution.}}
    \label{fig:execution_loop}
    % \vskip -0.2cm
\end{figure*}

Fig.~\ref{fig:execution_loop} describes the task planning and execution coupling. \revii{The goal is defined by set of goal conditions (predicates) \Goalcond.} We use the predicates $\predicate \in \Predicateset$ to predict a symbolic or logical state $l$ given an observation $\obs$ from the environment. The task planner finds a sequence of skills from the set of skills $\Skillset$ that along with their preconditions and effects define the planning domain. While in practice we use a simple planner from prior work~\cite{paxton2019representing}, any of a large number of task planners could be used, such as FastDownward~\cite{helmert2006fast}. \reviii{The logical effects \Effect for a skill are only used during task planning to update the logical state, and the corresponding predicates are never evaluated. The preconditions \Precond for a skill are evaluated during execution to verify if the skill can be executed at the current state.}

%We also define a task model which defines a list of high-level actions (e.g. \reachblock, \stack, etc.) along with a list of explicit preconditions and effects.
%The task planner operates on this symbolic state-space to generate a plan \plan to reach a goal state defined by a set of predicates \Goalcond.
%However, as described in~\cite{paxton2019representing}, we cannot simply execute any action whose explicit preconditions are met, since each operator can be sequenced in any order. Instead, we back-propagate preconditions from the goal state through the plan to enforce an ordering. Conditions are back-propagated until they were created by an action's effects. These extra implicit preconditions are combined with the action's explicit preconditions to form its complete preconditions set \Precond. 

% \subsection{Execution}
\begin{figure*}[tb]
    \centering
    \includegraphics[width=\textwidth]{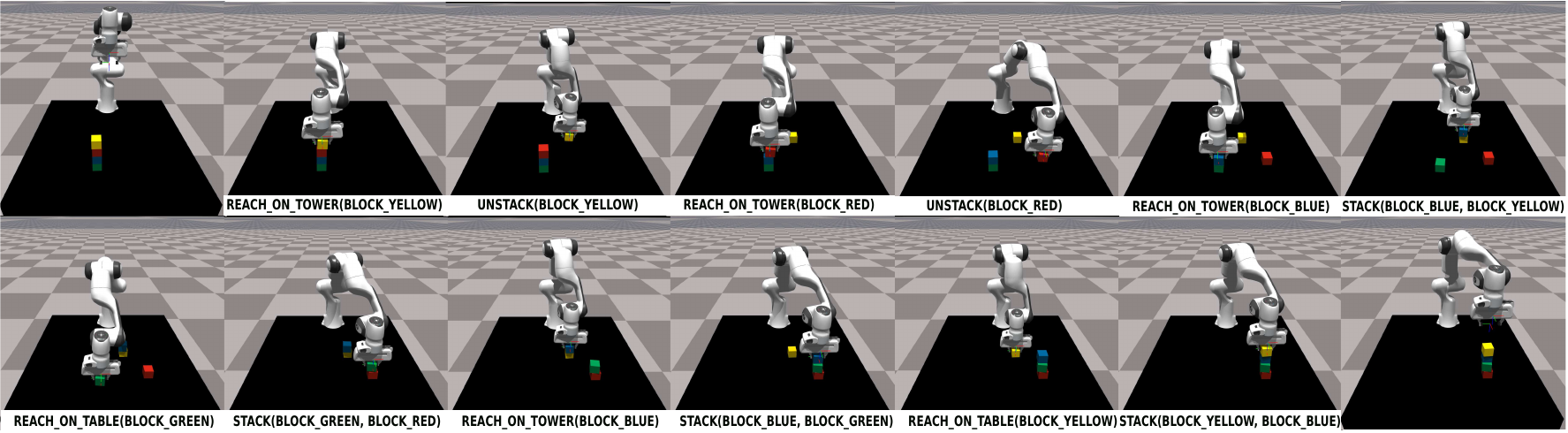}
    \caption{Example execution of a plan for reordering a tower of blocks: (green, blue, red, yellow) to (red, green, blue, yellow). The plan comprises a sequence of skills along with their arguments.}
    \label{fig:plan}
    % \vskip -0.5cm
\end{figure*}

The plan comprising a sequence of skills along with their arguments is executed with Alg.~\ref{alg:execution}. The preconditions allow the system to be reactive by determining if the preconditions of the skills in the plan $P$ are met, and switching skills as necessary. The algorithm executes the skills in the plan in sequence, starting from the first skill. If at any point, the preconditions for a skill to be executed are not met, the algorithm sequentially checks the preconditions for the previous skills in the plan, in reverse, till it comes across a skill for which all the preconditions are satisfied (Lines~\ref{alg:execution/find_skill_beg}-\ref{alg:execution/find_skill_end}). We keep track of the number of times each skill has been tried (Line~\ref{alg:execution/track_retrials}) and allow a maximum of \textsc{MaxRetrials} retrials for each skill. If the preconditions for none of the skills in the plan are satisfied or the retrials are expired for any skill, the algorithm replans (Lines~\ref{alg:execution/replan2},~\ref{alg:execution/replan3}~and~\ref{alg:execution/replan1}). We cap the number of times the algorithm replans to \textsc{MaxReplans}. Fig.~\ref{fig:plan} shows an example execution of a plan for reordering a stack of blocks. 

% \begin{figure}[!htb]
%     \centering
%     \includegraphics[width=1.0\columnwidth]{figures/plan.eps}
%     \caption{Example execution of a plan for reordering a tower of blocks.}
%     \label{fig:plan}
% \end{figure}

\begin{algorithm}[tb]
\caption{\label{alg:execution} Execution}
\begin{footnotesize}
\begin{algorithmic}[1]
\Procedure{Execute}{$\Goalcond$}
    \State replan\_counter $\gets 0$
    \While{replan\_counter $<$ \textsc{MaxReplans}}
        \State $\reviii{\obs \gets \textsc{Observe()}}$
        \State $\plan \gets \textsc{Plan}(\obs, \Goalcond)$ \label{alg:execution/replan1}
        \State replan\_counter $\mathrel{{+}{=}} 1$
        \If{$\textsc{ExecutePlan}(\obs, \Goalcond, \plan)$ = \textsc{Success}}
            \State\Return \textsc{Success}
        \EndIf
    \EndWhile
    \State\Return \textsc{Failure}
\EndProcedure
\Procedure{ExecutePlan}{$\obs, \Goalcond, \plan$}
    \State $i \gets 0$
    \While{$i < |\plan|$}
        \State $\skill \gets \plan_i $
        \While{$ \textbf{not } \forall \predicate \in \Precond^\skill,~\predicate(\obs) = \text{True}$} \Comment Pre-conditions check \label{alg:execution/find_skill_beg}
            \State $i \gets i - 1$
            \If{$i < 0$}
                \State\Return \textsc{Failure} \label{alg:execution/replan2}
            \EndIf
            \State $\skill \gets \plan_i$ \label{alg:execution/find_skill_end}
        \EndWhile
        \State retrial\_counter$[\skill] \mathrel{{+}{=}} 1$ \label{alg:execution/track_retrials} 
        % \IF{$retrial_counter[\skill] > \text{MAX_RETRIALS}$}
        \If{retrial\_counter$[\skill] >$ \textsc{MaxRetrials}}
            \State\Return \textsc{Failure} \label{alg:execution/replan3}
        \EndIf
        \State $\textsc{ExecuteSkill}(\skill, \obs)$ 
        \State $\reviii{\obs \gets \textsc{Observe()}}$
        % \State $\skill \gets \plan_{i+1} $
        \If{$\forall \predicate \in \Goalcond,~\predicate(\obs) = \text{True}$}  \Comment  Goal conditions check
            \State\Return \textsc{Success}
        \EndIf
        \State $i \gets i + 1$
    \EndWhile
    \State\Return \textsc{Failure}
\EndProcedure
\end{algorithmic}
\end{footnotesize}
\end{algorithm}

% \subsection{Lifelong learning}
% Though our models trained entirely in simulation transfer well to the real-world, we want our framework to continue to learn and improve the skills during real-world execution. We therefore fine-tune the models for the learned skills and the associated termination conditions with data collected during real-world execution. In order to use the real-world execution data, we define  \emph{action-level} success and \emph{task-level} success criteria for a skill trajectory during execution in the following manner. A skill trajectory is classified as successful or unsuccessful on an action-level basis if the preconditions of the next skill in the plan are met at the end of the skill. A skill trajectory is deemed successful on a task-level basis, if that trajectory is part of a successful episode.  

% The skill trajectories that are deemed successful on a task-level basis are used to fine-tune the skill models. All the states on a skill trajectory that is unsuccessful on an action-level basis along with all but the last state on a skill trajectory that is successful on an action-level basis, serve as negative examples for the corresponding termination model. Whereas, the last state on a skill trajectory that is successful on an action-level basis serves as a positive example.

% \FloatBarrier
\section{Evaluation}

\begin{table}[bt]
\resizebox{
\columnwidth}{!}{
\begin{tabular}{ccccc}
\toprule
& Replanning &  Retrials  & Successes/Failures    & Success Rate (\%) 
\\ \midrule
No retrials or replanning & No         &  No        & 229/21               & 91.6\%     \\
Retrials-only & No         &  Yes       &  239/11                  & 95.6\%      \\ 
\textbf{Retrials and replanning} & \textbf{Yes}        &  \textbf{Yes}       & \textbf{245/5}                 & \textbf{98.0\%}      \\ 
\bottomrule
\end{tabular}
}
\caption{\small{Success rate for stacking in simulation for 250 trials. The learned precondition models allow the system to recognize failures, and either retry the individual action or replan as necessary, greatly increasing success rate.}}
\label{tab:stack_success_sim}
\end{table}

\begin{table*}[bt]
\resizebox{\textwidth}{!}{%
\begin{tabular}{ccccclcccc}
\toprule
&&                      &                    &              &                     & \multicolumn{3}{c}{plan length of failures} & \\ 
& Reset every episode             & Replanning &   Retrials  & Success Rate (\%)    & \multicolumn{1}{c}{successes/failures} & 12 actions     & 10 actions    & 8 actions  & successful replans  \\ \midrule
No retrials or replanning             &yes        &     no               & no               & 84.0\%               & \multicolumn{1}{c}{210/40}             & 24          & 9            & 5       &                 -\\
Retrials-only             &yes        &     no               & yes               & 89.2\%               & \multicolumn{1}{c}{223/27}            & 14          & 8           & 4       &                 -\\
\textbf{Retrials and replanning}      &\textbf{yes}        &     \textbf{yes}              & \textbf{yes}              & \textbf{96.0\%}               & \multicolumn{1}{c}{\textbf{240/10}}             & \textbf{8}           & \textbf{1}            & \textbf{1}       &                 \textbf{22}\\\midrule
No retrials or replanning             &no         &     no               & no               & 83.2\%               & \multicolumn{1}{c}{208/42}             & 33          & 6            & 2       &                  -\\
Baseline 2: Retrial-only             &no         &     no               & yes               & 84.8\%               & \multicolumn{1}{c}{212/38}              & 23          & 11            & 4       &                  -\\
\textbf{Retrials and replanning}      &\textbf{no}         &     \textbf{yes}              & \textbf{yes}              & \textbf{93.2\%}               & \multicolumn{1}{c}{\textbf{232/18}}             & \textbf{12}          & \textbf{3}            & \textbf{3}       &                 \textbf{25}\\
\bottomrule
\end{tabular}%
}
\caption{\small{Success rate for reordering in simulation with reset for 250 trials. The first three rows correspond to the case when the scene was being reset after every episode; the last three rows correspond to the case when the scene was being reset only after a failure. Reactivity is crucial to achieving high success rates, and allows the system to  continue to operate independently with very few manual resets.}}
\label{tab:reorder_success_sim}
% \vskip -0.3cm
\end{table*}

We evaluate the proposed approach in simulation and in the real-world. Our goal is to test the core theory of our work: that the task-planning structure of our skills, with preconditions \Precond and effects \Effect, will be more robust in the real-world because these logical-state classifiers transfer well and allow our system to respond to failures.
We compare our approach to two baselines based on variations of the proposed approach:

\textbf{(1) No retrials or replanning}: The robot does not retry an action in the plan on failure and it does not replan. In other words, \textsc{MaxReplans} and \textsc{MaxRetrials} are both set to 0.

\textbf{(2) Retrials-only}: The robot can retry an action on failure, but it does not replan. This essentially only tests the preconditions in each skill's \Precond: we do not need to compute the entire logical state, hence cannot \emph{replan} to recover from a difficult situation.

We test on two tasks not seen during training: stacking and reordering. In the stacking task, the robot builds a tower of four blocks in a desired order, starting from the table. In the reordering task, we start with all four blocks stacked, and the robot must rearrange them in a completely different order. In the training data, we only saw individual skills, not the entire task, so both of these show the generalization ability of our proposed system.

Our system comprises a Franka Panda arm with a Microsoft Azure Kinect camera mounted to one side. We use PoseCNN~\cite{xiang2017posecnn} to generate the segmentation masks. Note that the pose estimates produced by PoseCNN were not accurate enough to complete the task due to range from the camera. For all experiments we use a set of four colored blocks (\blockred, \blockgreen, \blockblue, \blockyellow). In either case, we call an episode successful when the final desired order of the stack has been achieved.

\begin{figure}[tb]
    \centering
    \begin{subfigure}{.28\columnwidth}
        \centering
        \includegraphics[width=\columnwidth]{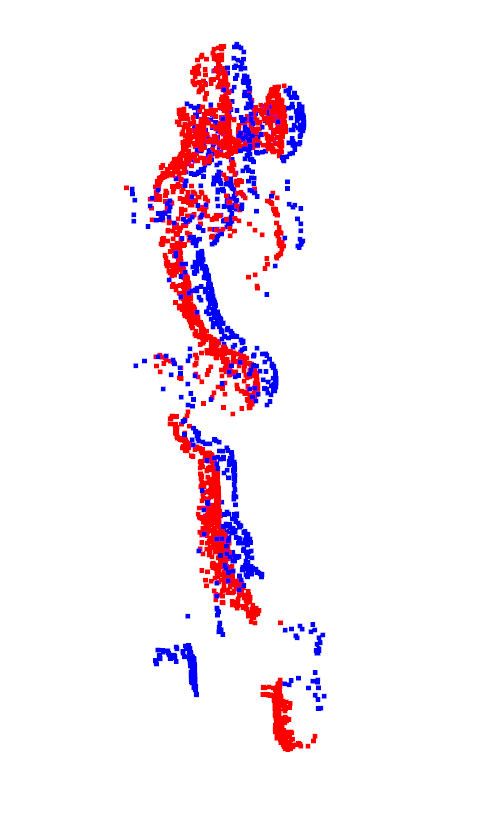}
        \caption{}
        \label{fig:sim_real_pc_front}
    \end{subfigure}
    \begin{subfigure}{.69\columnwidth}
        \centering
        \includegraphics[width=\columnwidth]{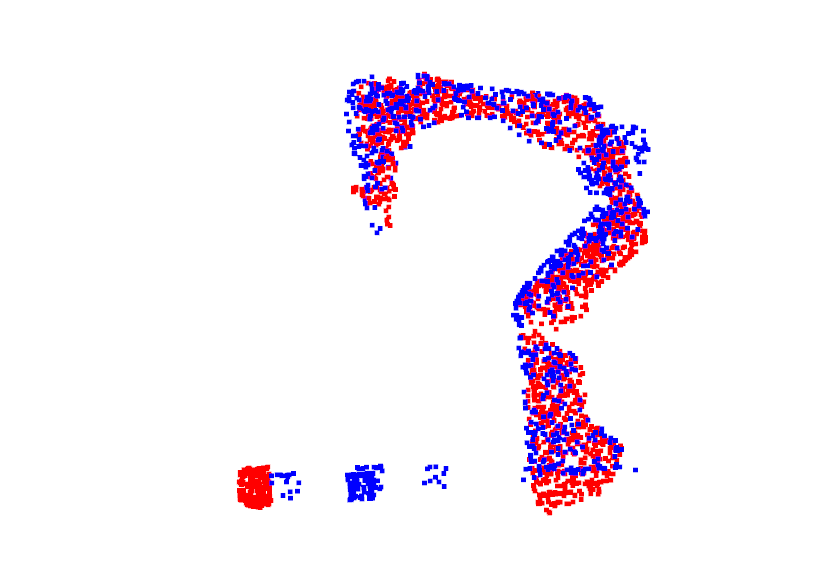}
        \caption{}
        \label{fig:sim_real_pc_side}
    \end{subfigure}\\
    \caption{We do not assume perfect calibration when training our models. (\subref{fig:sim_real_pc_front}) Front and (\subref{fig:sim_real_pc_side}) side view of the point cloud in simulation (blue) vs in the real-world (red). The mismatch can be handled with models that are trained with noise in the camera extrinsic parameters. Note that the blocks in simulation (blue) were not at the same poses as the blocks in the real-world (red).} %  and therefore the point clouds of the blocks are not indicative of the calibration mismatch.
    \label{fig:sim_real_pc}
    % \vskip -0.5cm
\end{figure}

\subsection{Implementation Details}

We took several additional steps to improve training and sim-to-real transfer, and had to set several tunable hyperparameters at runtime as described here.

%Following are some implementation details, training tricks and hyper-parameter decisions:  
% \begin{enumerate}
\textbf{Over-sampling points on objects of interest.} For faster training of our models, we sub-sample 2000 points from the camera point cloud. However this could lead to very few or no points on the objects of interest for the skill/predicates being trained. Therefore we reserve 10\% of points for the objects of interest (arguments). We also over-sample points on the objects of interest, if need be, in order to ensure that there are enough points from them in the point cloud.   

\textbf{Robustness to camera calibration.} Fig.~\ref{fig:sim_real_pc} compares the point clouds in simulation with the real-world. To make the models robust to the slight misalignment caused due to inaccurate camera calibration, we train them with random noise in the camera extrinsic parameters.

\textbf{Execution hyper-parameters.} We set both the \textsc{MaxReplans} and \textsc{MaxRetrials} parameters  in Alg.~\ref{alg:execution} to 5 for all experiments when using our approach. In practice these could be set higher to achieve even better performance.
 
\subsection{Simulation Results}

The results for stacking in simulation are shown in Table~\ref{tab:stack_success_sim}. We observe that with retrials and replanning using the predicate models, we achieve a very high success rate. Similarly, Table~\ref{tab:reorder_success_sim} shows the result of reordering in simulation. Compared to stacking, reordering typically involves longer plans which have a higher failure rate. Despite that, our framework achieves a $96\%$ success rate with retrials and replanning. In simulation, the most common cause of failure in our approach was the blocking of the camera's view of the other blocks by a partially constructed tower.

\subsection{Real-world Results}

In the real-world, we evaluate stacking for 10 trials. We observe that with no replanning and retrials, the robot succeeds in building a full tower of 4 blocks only once. It does a much better job when retrials are allowed. However there are cases when replanning becomes essential.
For example, if a partially-constructed tower topples in such a way that a block goes out of the robot's workspace, the robot needs to replan to figure out that it needs to execute \pull\xspace on that block to bring it back within its workspace.
Our full system with both retrials and replanning enabled succeeds 8/10 times with no adaption or fine-tuning. The remaining two times, it successfully constructs a partial tower of 3 blocks. In the real-world, the most common cause of failure was the partially-constructed tower toppling with some of the blocks going too far out of the robot's workspace such that \pull\xspace cannot be executed.
For videos, see the supplementary materials\footnote{Experiment videos: \url{https://www.youtube.com/playlist?list=PL-oD0xHUngeLfQmpngYkGFZarstfPOXqX}}.

\begin{table}[htp]
\resizebox{\columnwidth}{!}{%
\begin{tabular}{ccccccc}
\toprule
&                          &                          & \multicolumn{3}{c}{Size of successfully built towers} \\ 
                           &Replanning          &   Retrials               & 4 blocks         & 3 blocks         & 2 blocks         \\ \midrule
No retrials or replanning                  &No                  &   No                     & 1               & 3                 & 3 \\
Retrials-only                           &No                        &   Yes                    & 6                & 1                 & - \\
\textbf{Retrials and replanning}                           &\textbf{Yes}                 &   \textbf{Yes}                    & \textbf{8}                & \textbf{2}                & -         \\ 
\bottomrule
\end{tabular}%
}
\caption{\small{Stacking in the real-world for 10 trials. Our proposed approach succeeds 8/10 times. The remaining two times, it successfully constructs a tower of 3 blocks.}}
\label{tab:stack_success_real}
\end{table}

% \begin{table}[htp]
% \resizebox{\columnwidth}{!}{%
% \begin{tabular}{|c|c|c|c|c|}
% \toprule
% Task                & Total trials   & Successes       &  Failures/Resets           \\ \midrule
% Reordering          & 250            &  232            & 18                          \\
% \bottomrule
% \end{tabular}%
% }
% \caption{\small{Success rate for reordering without reset in sim.}}
% \label{tab:reorder_continuous_success_sim}
% \end{table}

% \begin{table*}[htp]
% %\resizebox{\columnwidth}{!}{%
% \begin{tabular}{|c|c|cl|ccccc|c|}
% \toprule
%           &                      &                      &                                         & \multicolumn{5}{c|}{Plan length of failures} & \\ 
% Replanning &    Retrials  & Success rate (\%)    & \multicolumn{1}{c|}{Successes/Failures(Resets)} & 12 actions     & 10 actions    & 8 actions & 6 actions & 2 actions  & Successful replans  \\ \midrule
% No         &     No               & -\%               & \multicolumn{1}{c|}{-/-}            &    -          & -            &    -       &  -   & - & -\\
% Yes        &     Yes              & 92.8\%               & \multicolumn{1}{c|}{232/18}             & 12             & 3            & 3    &  0    & 0    &  25\\
% \bottomrule
% \end{tabular}%
% %}
% \caption{\small{Success rate for reordering in simulation without reset for 250 trials with and without replanning. Reactivity is crucial to achieving high success rates, and allows the system to  continue to operate independently with very few manual resets.}}
% \label{tab:reorder_success_sim}
% \end{table*}

% \FloatBarrier
\section{Conclusions}
We proposed a framework for learning visuomotor skills from demonstration data collected in simulation that can be used to execute long-horizon multi-step manipulation tasks in the real-world.
%The framework uses learned predicate models for capturing the high-level symbolic state of the world and plans over this high-level state space.
Our system uses reactivity and replanning to adapt to failures. We tested on two scenarios: stacking and reordering, neither of which appeared in the training set.
%Using these predicate models, the system is able to react to failures during execution and recover by either retrying an action or by replanning.
Critically, our system successfully transfers to the real-world with no real-world demonstrations. \shohin{As mentioned in section \ref{sec:introduction}, execution in the real-world is not as precise as in simulation because of the reality gap. Specifically in our case the point clouds in the real-world are noisy unlike in simulation. Moreover, real-world execution is asynchronous leading to latency error i.e. there is a delay from when an observation is measured at the sensor to when the action is actually executed by the robot's actuators, which are not modelled by the simulator \cite{ibarz2021train}. Therefore fine-tuning the learned skills on real-world execution data can improve precision.} Because the system can recognize its own failures and successes, we can use it to collect its own training data for fine tuning on real-world data. In the future, we will use this system to develop lifelong learning for long-horizon tasks, where the system can collect large numbers of examples autonomously.
%We carry out comprehensive evaluation of our approach in both simulation and the real-world and show that our approach of reactive execution of skills using the predicate models outperforms naive baselines of open-loop execution of skills.  

% \FloatBarrier
%\section{Future Work}
%\input{07future_work}

% \FloatBarrier
\bibliographystyle{IEEEtran}
\bibliography{main}

\end{document}